\pdfoutput=1
\documentclass[11pt]{article}

\usepackage[]{EMNLP2022}

\usepackage{times}
\usepackage{latexsym}
\usepackage{graphicx}
\usepackage{xcolor}
\usepackage{colortbl}
\usepackage{paralist}
\usepackage{amsmath}
\usepackage{pifont}

\usepackage[T1]{fontenc}
\usepackage[utf8]{inputenc}

\usepackage{microtype}

\usepackage{inconsolata}

\title{VicunaNER: Zero/Few-shot Named Entity Recognition using Vicuna}

\author{Bin Ji\\
         National University of Singapore}
\begin{document}
\maketitle
\begin{abstract}

Large Language Models (LLMs, e.g., ChatGPT) have shown impressive zero- and few-shot capabilities in Named Entity Recognition (NER). However, these models can only be accessed via online APIs, which may cause data leak and non-reproducible problems.
In this paper, we propose VicunaNER, a zero/few-shot NER framework based on the newly released open-source LLM -- Vicuna. VicunaNER is a two-phase framework, where each phase leverages multi-turn dialogues with Vicuna to recognize entities from texts. We name the second phase as \textit{Re-Recognition}, which recognizes those entities not recognized in the first phase (a.k.a. \textit{Recongition}). 
Moreover, we set entity correctness check dialogues in each phase to filter out wrong entities.
We evaluate VicunaNER's zero-shot capacity on 10 datasets crossing 5 domains and few-shot capacity on Few-NERD. Experimental results demonstrate that VicunaNER achieves superior performance in both shot settings. Additionally, we conduct comprehensive investigations on Vicuna from multiple perspectives.

\end{abstract}

\section{Introduction}

Named Entity Recognition (NER) 
serves as a precondition for many downstream Natural Language Processing (NLP) tasks such as relation extraction. 
Deep supervised learning NER methods require extensive entity annotations, and it is hard to transfer them across domains. Zero- and few-shot NER is targeted in this scenario, which calls for zero or a few annotated examples and is capable of domain transferring.

Prototypical networks have been widely investigated for zero/few-shot NER, such as StructShot \cite{structshot}, CONTaiNER \cite{container}, ESD \cite{esd}, DecomMetaNER \cite{decommetaner}, and EP-Net \cite{epnet}. However, these networks still require fine-tuning datasets of thousands or tens of thousands of examples. 

\citet{gpt3} demonstrate that scaling up language models significantly improves task-agnostic, few-shot NLP task performance, and they propose GPT-3, the well-known milestone of Large Language Models (LLMs). GPT-3 achieves promising performance in diverse NLP tasks without any gradient updates or fine-tuning.  
Inspired by GPT-3, numerous LLMs are pre-trained or fine-tuned such as InstructGPT \cite{instructgpt}, Chinchilla \cite{chinchilla}, ChatGPT\footnote{https://chat.openai.com/chat}, PaLM \cite{palm} and GPT-4 \cite{gpt4}.  
Based on these LLMs, zero- and few-shot NER has been comprehensively investigated. 
For example, \citet{gpt3-biomedical} explore biomedical few-shot NER with GPT-3. And based on ChatGPT, \citet{gpt-icl} investigate document-level few-shot NER;
\citet{chatgpt-zeroshot} conduct research on zero-shot clinical NER; \citet{chatie} propose ChatIE to explore zero-shot information extraction including NER. Although these LLM-based studies achieve strong performance, and sometimes even reach competitiveness with prior best prototypical networks, the LLMs can only be accessed through online APIs, which causes the following problems:
\begin{compactitem}
\item[1.] Data leak problem. For example, sensitive data from Samsung was leaked to ChatGPT.\footnote{https://techcrunch.com/2023/05/02/samsung-bans-use-of-generative-ai-tools-like-chatgpt-after-april-internal-data-leak/}
\item[2.] Non-reproducible problem. Because the LLMs are fine-tuned constantly, but the details are not publicly available \cite{chatlog}.
\end{compactitem}

Fortunately, some open-source LLMs are available to the public, such as T5 \cite{t5}, OPT \cite{opt}, GLM \cite{glm}, BLOOM \cite{bloom}, and LLaMA \cite{llama}.
Especially, LLaMA attracts much research attention due to that: (1) it can be deployed on local servers; (2) it has evolved many powerful variants via fine-tuning, such as Alpaca \cite{alpaca}, Baize \cite{baize}, Koala \cite{koala}, and Vicuna \cite{vicuna}. 

With the goal of exploring unlimited zero- and few-shot NER approaches, we propose VicunaNER, a Vicuna-based framework that can conduct both zero- and few-shot NER. 
VicunaNER is composed of two phases, which are known as \textit{Recognition} and \textit{Re-Recogniziton}, respectively. 
\begin{compactitem}
\item[1.]
\textit{Recognition} consists of multi-turn dialogues with Vicuna. The first-turn prompts Vicuna to recognize entities from texts. 
For each of the recognized entities, we use one-turn dialogue to prompt Vicuna to check its correctness.
After doing this, \textit{Recognition} generates a list of entities for each text.\footnote{It is also possible that no entity is recognized.}
However, we observe that \textit{Recognition} fails to recognize numerous entities when analyzing the entity results, which motivates us to add the \textit{Re-Recogniziton} phase. 
\item[2.]
\textit{Re-Recogniziton} is also composed of multi-turn dialogues with Vicuna.
Given a text and its entities recognized in \textit{Recogniziton}, Vicuna is prompted to recognize those unrecognized entities in the first-turn dialogue. Then Vicuna is prompted to check the correctness of newly recognized entities in the other turn dialogues.
\end{compactitem}

Entities recognized in the two phases are merged as the NER results.  

We evaluate VicunaNER's zero-shot capacity on 10 datasets crossing 5 domains and few-shot capacity on Few-NERD \cite{few-nerd}. 
Experimental results show that: 
(1) Under the zero-shot setting, VicunaNER outperforms the ChatGPT-based ChatIE on xxx out of the xxx datasets, even ChatGPT is more powerful than Vicuna;  
(2) Under the few-shot setting, VicunaNER consistently surpasses the listed baselines, including LLM-based frameworks and prototypical networks.
Additionally, we conduct comprehensive investigations to disclose the drawbacks of Vicuna, providing references for fine-tuning it in the future.

\begin{figure*}[t]
\centering
\includegraphics[width=0.9\textwidth]{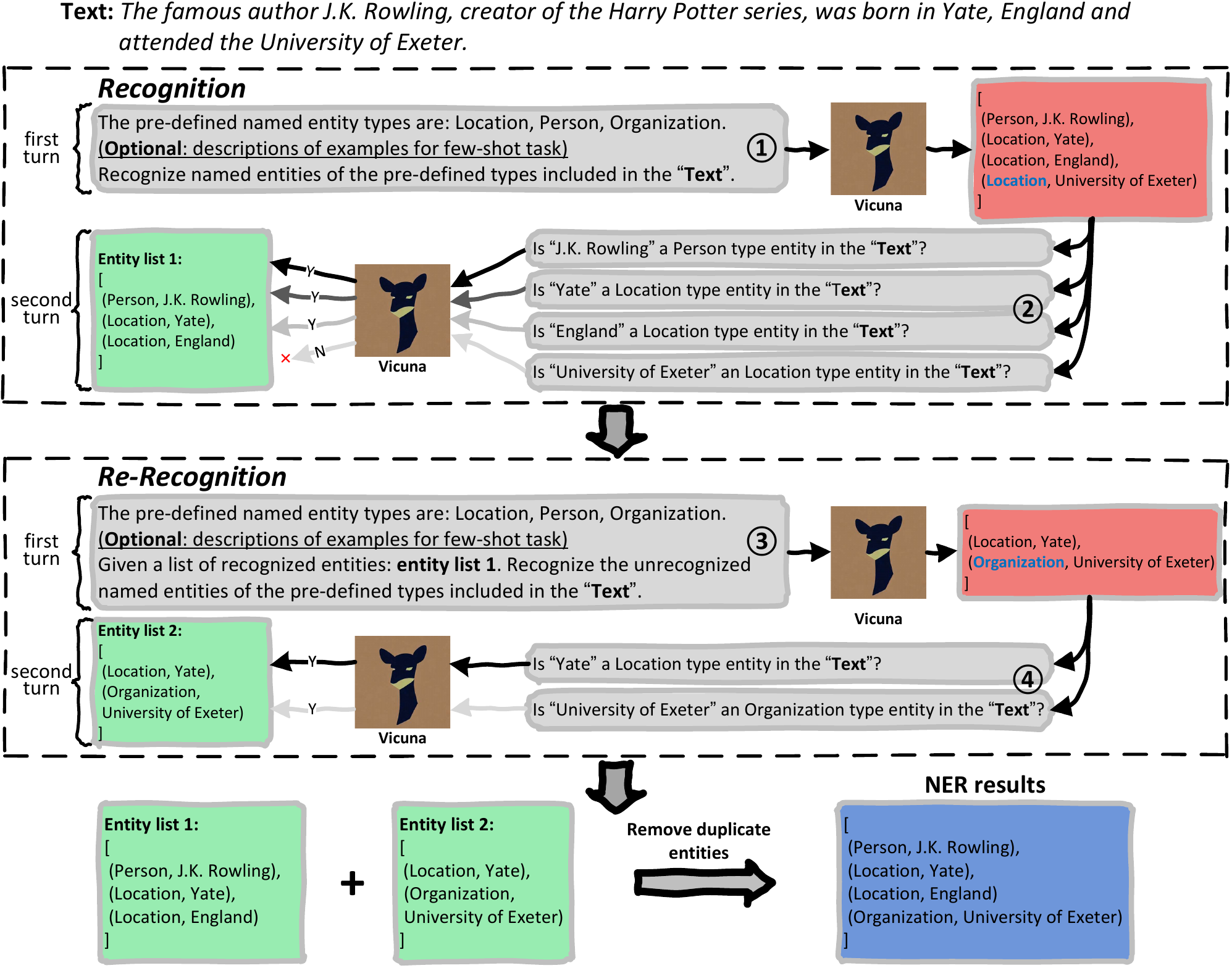} 
\caption{The architecture of VicunaNER. It is composed of two phases namely \textit{Recognition} and \textit{Re-Recognition}, and each phase consists of multi-turn dialogues with Vicuna. We use a zero-shot NER example to describe the workflow. {Texts in the gray background are prompts;} {entity lists in the red background manage entities recognized by the first-turn dialogue in each phase;}{entity lists in the green background manage entities recognized in each phase;} {the entity list in the blue background manages the entities recognized by VicunaNER. }}\label{figure1}
\label{model2}
\end{figure*}

\section{Related Work}

\section{VicunaNER}

As shown in Figure \ref{figure1}, Vicuna is composed of two phases, which are named as  \textit{Recognition} and \textit{Re-Recognition}, respectively. \textit{Recognition} is stacked upon \textit{Re-Recognition}, and both of them consist of multi-turn dialogues with Vicuna. \textit{Recogniton} conducts the first round of NER, and \textit{Re-Recognition} conducts another round of NER. The reasons for this stacked design are summarized as follows:
\begin{compactitem}
\item[1.] We find that Vicuna fails to recognize numerous entities in \textit{Recognition}. Hence the main purpose of \textit{Re-Recognition} is to recognize those unrecognized entities, which guarantees better model performance.
\item[2.] Different from the LLMs (e.g., ChatGPT) that only be allowed to access via online APIs, we can deploy the open-source Vicuna on a local server, which enables us to leverage Vicuna's generation capabilities without restrictions.\footnote{Note that Vicuna is intended for non-commercial use only.}
\end{compactitem} 

We will illustrate the architecture in $\S$ \ref{section-recognition} -- \ref{section-entitymerging}. Moreover, we will discuss more details in $\S$ \ref{section-discussion}.

\subsection{Recognition}\label{section-recognition}

\textit{Recognition} consists of multi-turn dialogues with Vicuna. 
Given a text and pre-defined named entity types, the first-turn dialogue prompts Vicuna to recognize entities included in the given text. 
For the zero-shot NER task, we combine descriptions of entity types and the given text to obtain a prompt; for the few-shot NER task, we combine descriptions of entity types, support examples, and the given text to obtain a prompt.
Next, we feed the prompt to Vicuna. 

We find there are several kinds of entity prediction errors when analyzing the recognized entities, including entity boundary error, entity type error, and non-entity text spans that are mistakenly recognized as entities. Figure \ref{figure1} shows an entity type error, ``University of Exeter'', which should be predicted as an Organization entity, is predicted as a Location entity actually. We use the other turn dialogues to filter these mistakenly recognized entities.
To be specific, for each predicted entity, we combine its text span, its type, and the text it is included to obtain a prompt. 
Figure \ref{figure1}-\ding{173} shows four prompt examples. 
Then we feed the prompt to Vicuna. 
For example, we filter out the ``(Location, University of Exeter)'' in Figure \ref{figure1}.
At last, we use a list to manage the remaining entities, as the ``\textbf{Entity list} 1'' in Figure \ref{figure1} shows.

For better comprehension, we report several real-world prompt examples of this phase in Appendix \ref{prompt-examples}.

\subsection{Re-Recognition}\label{section-rerecognition}
Actually, \textit{Recognition} achieves a whole round of zero/few-shot NER and we can terminate the NER process after obtaining the entity list. However, we find that Vicuna fails to recognize numerous entities when analyzing the entities recognized in \textit{Recognition}. Hence, we design the \textit{Re-Recognition} phase to recognize those unrecognized entities.

As shown in Figure \ref{figure1}, \textit{Re-Recognition} consists of multi-turn dialogues with Vicuna, which is similar to \textit{Recognition}. The only difference is the prompts used in the first-turn dialogue of the two phases. To be specific, we add entity descriptions to the prompt used in this phase, where these entities are recognized in \textit{Recognition}, as Figure \ref{figure1}-\ding{174} shows. The purpose of doing this is to guide Vicuna in recognizing those unrecognized entities solely. We also use a list to manage entities recognized in \textit{Re-Recognition}, as the ``\textbf{Entity list 2}'' in Figure \ref{figure1} shows. 

Although the prompt is designed to ask Vicuna to solely recognize those unrecognized entities, we find that Vicuna still recognizes some already recognized entities, such as the ``(Location, Yate)'' shown in Figure \ref{figure1}. We attribute it to the fact that Vicuna has limitations in ensuring the factual accuracy of its outputs \cite{vicuna}.

For better comprehension, we report real-world prompt examples of this phase in Appendix \ref{prompt-examples}.

\subsection{Entity Merging}\label{section-entitymerging}
As aforementioned, we obtain one entity list in each of the two phases, but there may be entities that overlap between the two entity lists. Hence, we remove these overlapping entities when merging the two lists to obtain the NER results, as shown in Figure \ref{figure1}.

\subsection{Discussion}\label{section-discussion}
\subsubsection{Comparison of VicunaNER and ChatIE}
Concurrent with our work, ChatIE is ChatGPT-based framework that can conduct zero-shot NER, and it also adopts a two-phase architecture. We claim that our VicunaNER is quite different from ChatIE in the following aspects:
\begin{compactitem}
\item[1.] Our VicunaNER depends on the open-source Vicuna, while ChatIE is built upon the more powerful but restricted ChatGPT API.
\item[2.] Our VicunaNER conducts a whole round of NER in each of its two phases,
While ChatIE solely extracts entity types in its first phase and recognizes entities according to the extracted types in its second phase. 
\item[3.] Our VicunaNER can conduct both zero- and few-shot NER tasks, while ChatIE is only designed to perform the Zero-shot NER task. 
\end{compactitem}

\subsubsection{Are More Re-Recognition Phases Necessary?}

It seems that adding more \textit{Re-Recognition} phases can trigger better zero/few-shot NER performance. However, we demonstrate that adding more than one \textit{Re-Recognition} phase solely brings tiny performance improvements but greatly increases model inference time. We conduct experimental investigations on counts of the \textit{Re-Recognition} phase in $\S$ {xxx}

\subsubsection{Entity Form}

Following the established line of work \cite{chatie}, we don't prompt Vicuna to output entity locations because it is hard for LLMs to output the exact locations. This may cause confusion when an entity occurs more than once in a given text but VicunaNER only recognizes some of them.

\bibliography{anthology,custom}

\begin{thebibliography}{25}
\expandafter\ifx\csname natexlab\endcsname\relax\def\natexlab#1{#1}\fi

\bibitem[{Brown et~al.(2020)Brown, Mann, Ryder, Subbiah, Kaplan, Dhariwal,
  Neelakantan, Shyam, Sastry, Askell, Agarwal, Herbert-Voss, Krueger, Henighan,
  Child, Ramesh, Ziegler, Wu, Winter, Hesse, Chen, Sigler, Litwin, Gray, Chess,
  Clark, Berner, McCandlish, Radford, Sutskever, and Amodei}]{gpt3}
Tom Brown, Benjamin Mann, Nick Ryder, Melanie Subbiah, Jared~D Kaplan, Prafulla
  Dhariwal, Arvind Neelakantan, Pranav Shyam, Girish Sastry, Amanda Askell,
  Sandhini Agarwal, Ariel Herbert-Voss, Gretchen Krueger, Tom Henighan, Rewon
  Child, Aditya Ramesh, Daniel Ziegler, Jeffrey Wu, Clemens Winter, Chris
  Hesse, Mark Chen, Eric Sigler, Mateusz Litwin, Scott Gray, Benjamin Chess,
  Jack Clark, Christopher Berner, Sam McCandlish, Alec Radford, Ilya Sutskever,
  and Dario Amodei. 2020.
\newblock \href
  {https://proceedings.neurips.cc/paper_files/paper/2020/file/1457c0d6bfcb4967418bfb8ac142f64a-Paper.pdf}
  {Language models are few-shot learners}.
\newblock In \emph{Advances in Neural Information Processing Systems},
  volume~33, pages 1877--1901. Curran Associates, Inc.

\bibitem[{Chiang et~al.(2023)Chiang, Li, Lin, Sheng, Wu, Zhang, Zheng, Zhuang,
  Zhuang, Gonzalez, Stoica, and Xing}]{vicuna}
Wei-Lin Chiang, Zhuohan Li, Zi~Lin, Ying Sheng, Zhanghao Wu, Hao Zhang, Lianmin
  Zheng, Siyuan Zhuang, Yonghao Zhuang, Joseph~E. Gonzalez, Ion Stoica, and
  Eric~P. Xing. 2023.
\newblock \href {https://vicuna.lmsys.org} {Vicuna: An open-source chatbot
  impressing gpt-4 with 90\%* chatgpt quality}.

\bibitem[{Das et~al.(2022)Das, Katiyar, Passonneau, and Zhang}]{container}
Sarkar Snigdha~Sarathi Das, Arzoo Katiyar, Rebecca Passonneau, and Rui Zhang.
  2022.
\newblock \href {https://doi.org/10.18653/v1/2022.acl-long.439} {{CONT}ai{NER}:
  Few-shot named entity recognition via contrastive learning}.
\newblock In \emph{Proceedings of the 60th Annual Meeting of the Association
  for Computational Linguistics (Volume 1: Long Papers)}, pages 6338--6353,
  Dublin, Ireland. Association for Computational Linguistics.

\bibitem[{Ding et~al.(2021)Ding, Xu, Chen, Wang, Han, Xie, Zheng, and
  Liu}]{few-nerd}
Ning Ding, Guangwei Xu, Yulin Chen, Xiaobin Wang, Xu~Han, Pengjun Xie, Haitao
  Zheng, and Zhiyuan Liu. 2021.
\newblock \href {https://doi.org/10.18653/v1/2021.acl-long.248} {Few-{NERD}: A
  few-shot named entity recognition dataset}.
\newblock In \emph{Proceedings of the 59th Annual Meeting of the Association
  for Computational Linguistics and the 11th International Joint Conference on
  Natural Language Processing (Volume 1: Long Papers)}, pages 3198--3213,
  Online. Association for Computational Linguistics.

\bibitem[{Driess et~al.(2023)Driess, Xia, Sajjadi, Lynch, Chowdhery, Ichter,
  Wahid, Tompson, Vuong, Yu, Huang, Chebotar, Sermanet, Duckworth, Levine,
  Vanhoucke, Hausman, Toussaint, Greff, Zeng, Mordatch, and Florence}]{palm}
Danny Driess, Fei Xia, Mehdi S.~M. Sajjadi, Corey Lynch, Aakanksha Chowdhery,
  Brian Ichter, Ayzaan Wahid, Jonathan Tompson, Quan Vuong, Tianhe Yu, Wenlong
  Huang, Yevgen Chebotar, Pierre Sermanet, Daniel Duckworth, Sergey Levine,
  Vincent Vanhoucke, Karol Hausman, Marc Toussaint, Klaus Greff, Andy Zeng,
  Igor Mordatch, and Pete Florence. 2023.
\newblock \href {http://arxiv.org/abs/2303.03378} {Palm-e: An embodied
  multimodal language model}.

\bibitem[{Geng et~al.(2023)Geng, Gudibande, Liu, Wallace, Abbeel, Levine, and
  Song}]{koala}
Xinyang Geng, Arnav Gudibande, Hao Liu, Eric Wallace, Pieter Abbeel, Sergey
  Levine, and Dawn Song. 2023.
\newblock \href {https://bair.berkeley.edu/blog/2023/04/03/koala/} {Koala: A
  dialogue model for academic research}.
\newblock Blog post.

\bibitem[{He et~al.(2023)He, Wang, Hu, Liu, Liu, Xu, and Shen}]{gpt-icl}
Jiabang He, Lei Wang, Yi~Hu, Ning Liu, Hui Liu, Xing Xu, and Heng~Tao Shen.
  2023.
\newblock \href {http://arxiv.org/abs/2303.05063} {Icl-d3ie: In-context
  learning with diverse demonstrations updating for document information
  extraction}.

\bibitem[{Hoffmann et~al.(2022)Hoffmann, Borgeaud, Mensch, Buchatskaya, Cai,
  Rutherford, de~las Casas, Hendricks, Welbl, Clark, Hennigan, Noland,
  Millican, van~den Driessche, Damoc, Guy, Osindero, Simonyan, Elsen, Vinyals,
  Rae, and Sifre}]{chinchilla}
Jordan Hoffmann, Sebastian Borgeaud, Arthur Mensch, Elena Buchatskaya, Trevor
  Cai, Eliza Rutherford, Diego de~las Casas, Lisa~Anne Hendricks, Johannes
  Welbl, Aidan Clark, Tom Hennigan, Eric Noland, Katherine Millican, George
  van~den Driessche, Bogdan Damoc, Aurelia Guy, Simon Osindero, Karen Simonyan,
  Erich Elsen, Oriol Vinyals, Jack~William Rae, and Laurent Sifre. 2022.
\newblock \href {https://openreview.net/forum?id=iBBcRUlOAPR} {An empirical
  analysis of compute-optimal large language model training}.
\newblock In \emph{Advances in Neural Information Processing Systems}.

\bibitem[{Hu et~al.(2023)Hu, Ameer, Zuo, Peng, Zhou, Li, Li, Li, Jiang, and
  Xu}]{chatgpt-zeroshot}
Yan Hu, Iqra Ameer, Xu~Zuo, Xueqing Peng, Yujia Zhou, Zehan Li, Yiming Li,
  Jianfu Li, Xiaoqian Jiang, and Hua Xu. 2023.
\newblock \href {http://arxiv.org/abs/2303.16416} {Zero-shot clinical entity
  recognition using chatgpt}.

\bibitem[{Ji et~al.(2022)Ji, Li, Gan, Yu, Ma, Liu, and Yang}]{epnet}
Bin Ji, Shasha Li, Shaoduo Gan, Jie Yu, Jun Ma, Huijun Liu, and Jing Yang.
  2022.
\newblock \href {https://aclanthology.org/2022.coling-1.159} {Few-shot named
  entity recognition with entity-level prototypical network enhanced by
  dispersedly distributed prototypes}.
\newblock In \emph{Proceedings of the 29th International Conference on
  Computational Linguistics}, pages 1842--1854, Gyeongju, Republic of Korea.
  International Committee on Computational Linguistics.

\bibitem[{Jimenez~Gutierrez et~al.(2022)Jimenez~Gutierrez, McNeal, Washington,
  Chen, Li, Sun, and Su}]{gpt3-biomedical}
Bernal Jimenez~Gutierrez, Nikolas McNeal, Clayton Washington, You Chen, Lang
  Li, Huan Sun, and Yu~Su. 2022.
\newblock \href {https://aclanthology.org/2022.findings-emnlp.329} {Thinking
  about {GPT}-3 in-context learning for biomedical {IE}? think again}.
\newblock In \emph{Findings of the Association for Computational Linguistics:
  EMNLP 2022}, pages 4497--4512, Abu Dhabi, United Arab Emirates. Association
  for Computational Linguistics.

\bibitem[{Ma et~al.(2022)Ma, Jiang, Wu, Zhao, and Lin}]{decommetaner}
Tingting Ma, Huiqiang Jiang, Qianhui Wu, Tiejun Zhao, and Chin-Yew Lin. 2022.
\newblock \href {https://doi.org/10.18653/v1/2022.findings-acl.124} {Decomposed
  meta-learning for few-shot named entity recognition}.
\newblock In \emph{Findings of the Association for Computational Linguistics:
  ACL 2022}, pages 1584--1596, Dublin, Ireland. Association for Computational
  Linguistics.

\bibitem[{OpenAI(2023)}]{gpt4}
OpenAI. 2023.
\newblock \href {http://arxiv.org/abs/2303.08774} {Gpt-4 technical report}.

\bibitem[{Ouyang et~al.(2022)Ouyang, Wu, Jiang, Almeida, Wainwright, Mishkin,
  Zhang, Agarwal, Slama, Gray, Schulman, Hilton, Kelton, Miller, Simens,
  Askell, Welinder, Christiano, Leike, and Lowe}]{instructgpt}
Long Ouyang, Jeffrey Wu, Xu~Jiang, Diogo Almeida, Carroll Wainwright, Pamela
  Mishkin, Chong Zhang, Sandhini Agarwal, Katarina Slama, Alex Gray, John
  Schulman, Jacob Hilton, Fraser Kelton, Luke Miller, Maddie Simens, Amanda
  Askell, Peter Welinder, Paul Christiano, Jan Leike, and Ryan Lowe. 2022.
\newblock \href {https://openreview.net/forum?id=TG8KACxEON} {Training language
  models to follow instructions with human feedback}.
\newblock In \emph{Advances in Neural Information Processing Systems}.

\bibitem[{Raffel et~al.(2020)Raffel, Shazeer, Roberts, Lee, Narang, Matena,
  Zhou, Li, and Liu}]{t5}
Colin Raffel, Noam Shazeer, Adam Roberts, Katherine Lee, Sharan Narang, Michael
  Matena, Yanqi Zhou, Wei Li, and Peter~J. Liu. 2020.
\newblock \href {http://jmlr.org/papers/v21/20-074.html} {Exploring the limits
  of transfer learning with a unified text-to-text transformer}.
\newblock \emph{Journal of Machine Learning Research}, 21(140):1--67.

\bibitem[{Taori et~al.(2023)Taori, Gulrajani, Zhang, Dubois, Li, Guestrin,
  Liang, and Hashimoto}]{alpaca}
Rohan Taori, Ishaan Gulrajani, Tianyi Zhang, Yann Dubois, Xuechen Li, Carlos
  Guestrin, Percy Liang, and Tatsunori~B. Hashimoto. 2023.
\newblock Stanford alpaca: An instruction-following llama model.
\newblock \url{https://github.com/tatsu-lab/stanford_alpaca}.

\bibitem[{Touvron et~al.(2023)Touvron, Lavril, Izacard, Martinet, Lachaux,
  Lacroix, Rozière, Goyal, Hambro, Azhar, Rodriguez, Joulin, Grave, and
  Lample}]{llama}
Hugo Touvron, Thibaut Lavril, Gautier Izacard, Xavier Martinet, Marie-Anne
  Lachaux, Timothée Lacroix, Baptiste Rozière, Naman Goyal, Eric Hambro,
  Faisal Azhar, Aurelien Rodriguez, Armand Joulin, Edouard Grave, and Guillaume
  Lample. 2023.
\newblock \href {http://arxiv.org/abs/2302.13971} {Llama: Open and efficient
  foundation language models}.

\bibitem[{Tu et~al.(2023)Tu, Li, Yu, Wang, Hou, and Li}]{chatlog}
Shangqing Tu, Chunyang Li, Jifan Yu, Xiaozhi Wang, Lei Hou, and Juanzi Li.
  2023.
\newblock \href {http://arxiv.org/abs/2304.14106} {Chatlog: Recording and
  analyzing chatgpt across time}.

\bibitem[{Wang et~al.(2022)Wang, Xu, Liu, Zhou, Cao, Chang, and Sui}]{esd}
Peiyi Wang, Runxin Xu, Tianyu Liu, Qingyu Zhou, Yunbo Cao, Baobao Chang, and
  Zhifang Sui. 2022.
\newblock \href {https://doi.org/10.18653/v1/2022.naacl-main.369} {An enhanced
  span-based decomposition method for few-shot sequence labeling}.
\newblock In \emph{Proceedings of the 2022 Conference of the North American
  Chapter of the Association for Computational Linguistics: Human Language
  Technologies}, pages 5012--5024, Seattle, United States. Association for
  Computational Linguistics.

\bibitem[{Wei et~al.(2023)Wei, Cui, Cheng, Wang, Zhang, Huang, Xie, Xu, Chen,
  Zhang et~al.}]{chatie}
Xiang Wei, Xingyu Cui, Ning Cheng, Xiaobin Wang, Xin Zhang, Shen Huang, Pengjun
  Xie, Jinan Xu, Yufeng Chen, Meishan Zhang, et~al. 2023.
\newblock Zero-shot information extraction via chatting with chatgpt.
\newblock \emph{arXiv preprint arXiv:2302.10205}.

\bibitem[{Workshop et~al.(2023)Workshop, :, Scao, Fan, and et~al.}]{bloom}
BigScience Workshop, :, Teven~Le Scao, Angela Fan, and Christopher~Akiki et~al.
  2023.
\newblock \href {http://arxiv.org/abs/2211.05100} {Bloom: A 176b-parameter
  open-access multilingual language model}.

\bibitem[{Xu et~al.(2023)Xu, Guo, Duan, and McAuley}]{baize}
Canwen Xu, Daya Guo, Nan Duan, and Julian McAuley. 2023.
\newblock \href {http://arxiv.org/abs/2304.01196} {Baize: An open-source chat
  model with parameter-efficient tuning on self-chat data}.

\bibitem[{Yang and Katiyar(2020)}]{structshot}
Yi~Yang and Arzoo Katiyar. 2020.
\newblock \href {https://doi.org/10.18653/v1/2020.emnlp-main.516} {Simple and
  effective few-shot named entity recognition with structured nearest neighbor
  learning}.
\newblock In \emph{Proceedings of the 2020 Conference on Empirical Methods in
  Natural Language Processing (EMNLP)}, pages 6365--6375, Online. Association
  for Computational Linguistics.

\bibitem[{Zeng et~al.(2023)Zeng, Liu, Du, Wang, Lai, Ding, Yang, Xu, Zheng,
  Xia, Tam, Ma, Xue, Zhai, Chen, Liu, Zhang, Dong, and Tang}]{glm}
Aohan Zeng, Xiao Liu, Zhengxiao Du, Zihan Wang, Hanyu Lai, Ming Ding, Zhuoyi
  Yang, Yifan Xu, Wendi Zheng, Xiao Xia, Weng~Lam Tam, Zixuan Ma, Yufei Xue,
  Jidong Zhai, Wenguang Chen, Zhiyuan Liu, Peng Zhang, Yuxiao Dong, and Jie
  Tang. 2023.
\newblock \href {https://openreview.net/forum?id=-Aw0rrrPUF} {{GLM}-130b: An
  open bilingual pre-trained model}.
\newblock In \emph{The Eleventh International Conference on Learning
  Representations}.

\bibitem[{Zhang et~al.(2022)Zhang, Roller, Goyal, Artetxe, Chen, Chen, Dewan,
  Diab, Li, Lin, Mihaylov, Ott, Shleifer, Shuster, Simig, Koura, Sridhar, Wang,
  and Zettlemoyer}]{opt}
Susan Zhang, Stephen Roller, Naman Goyal, Mikel Artetxe, Moya Chen, Shuohui
  Chen, Christopher Dewan, Mona Diab, Xian Li, Xi~Victoria Lin, Todor Mihaylov,
  Myle Ott, Sam Shleifer, Kurt Shuster, Daniel Simig, Punit~Singh Koura, Anjali
  Sridhar, Tianlu Wang, and Luke Zettlemoyer. 2022.
\newblock \href {http://arxiv.org/abs/2205.01068} {Opt: Open pre-trained
  transformer language models}.

\end{thebibliography}
\bibliographystyle{acl_natbib}

\appendix

\section{Prompt Examples}
\label{prompt-examples}

This is a section in the appendix.
use the full name of named entity types rather than short names.

\end{document}